\documentclass[11pt]{article}

\usepackage[preprint]{acl}

\usepackage{times}
\usepackage{latexsym}

\usepackage[T1]{fontenc}

\usepackage[utf8]{inputenc}

\usepackage{microtype}

\usepackage{inconsolata}

\usepackage{xcolor}
\usepackage{graphicx}

\definecolor{c_persona}{RGB}{38, 166, 154}
\definecolor{c_instruction}{RGB}{239, 83, 80}
\definecolor{c_question}{RGB}{92, 107, 192}
\definecolor{c_option_symbol}{RGB}{255, 167, 38}

\usepackage{amsmath}
\usepackage{amssymb}

\newcommand{\COMSQ}{$\,\blacksquare\;$}

\newcommand{\Persona}{\textcolor{c_persona}{\COMSQ}}
\newcommand{\Instruction}{\textcolor{c_instruction}{\COMSQ}}
\newcommand{\Question}{\textcolor{c_question}{\COMSQ}}
\newcommand{\OptionSymbol}{\textcolor{c_option_symbol}{\COMSQ}}

\newcommand{\HYPRCBE}{\Persona\Instruction\Question\OptionSymbol}

\title{The Unsampled Truth: Quantifying Prompt Artifacts in LM Psychometrics}

\author{
  \textbf{Nils Schwager}, \textbf{Christoph Hau}, \textbf{Simon Münker}, \textbf{Achim Rettinger} \\
  Trier University, Trier, Germany \\
  \texttt{\{schwager, hau, muenker, rettinger\}@uni-trier.de} \\
}

\begin{document}
\maketitle

\begin{abstract}
When prompting language models for psychometric assessment, researchers assume that the responses reflect the injected persona and the meaning of the survey item. We test this premise using a diagnostic design that crosses five semantically distinct baseline personas with five semantically equivalent variants of each of four prompt components (persona wording, task instruction, item wording, option symbol). Measuring the 1-Wasserstein distance between the resulting response distributions and partitioning the variation among the five components allows for the separation of target effects from prompt artifacts. We apply the framework to 13 open-weight small language models (0.6B–14B) on the Big Five Inventory and the Short Dark Triad.  We find that in most models, the task instruction and option symbol displace response distributions further than paraphrasing the persona description or the item itself. For a substantial share of items, the artifact share of explained variation exceeds 50\%; non-semantic changes of the prompt account for more response variation than the baseline personas. Our framework lets researchers quantify these prompt artifacts before interpreting psychometric output.
\end{abstract}

\section{Introduction}
\label{sec:introduction}

When researchers prompt a language model (LM) to take a personality test, they are operating under a fragile premise. The current methodology involves feeding models specific personas and standardized inventories to extract simulated human attitudes and behaviors \cite{jiang2024personallm, lee2025llms, zheng2025lmlpa}. However, this paradigm rests on a critical assumption: that the model's predicted output is primarily a reflection of the injected persona and the semantic meaning of the survey item.

Ongoing research questions whether LMs can accurately predict human response behavior \cite{dominguez2024questioning, rupprecht2025prompt, pmlr-v202-santurkar23a, tjuatja2024llms, ahnert2025survey} or whether classical psychological principles, such as construct validity, can be applied to LMs \cite{munker-2025-fingerprinting}. These concerns are compounded by the models' sensitivity to prompt design. Studies demonstrate that altering components, such as answer option symbols \cite{yang2025option, 10.1371/journal.pone.0319159}, personas \cite{lutz2025prompt}, output constraints in the instruction \cite{rottger2024political}, and item reformulations \cite{tosato2026persistent}, induces substantial response variance.

However, core elements including the task instruction remain under-researched. More importantly, the field lacks a unified experimental design that evaluates prompt components (task instruction, option symbols, item \& persona wording) simultaneously. Because prior studies have not quantified the impact of variations of these components when their semantic meaning is held constant, the share of prompt artifacts\footnote{In classical psychometrics, an artifact is a finding that is produced artificially, for example by the measurement instrument itself. Accordingly, we define a prompt artifact as the distributional shift induced by semantically equivalent prompt variants, representing method variance rather than target effects \cite{podsakoff2003common}.} in LM psychometric predictions remains unknown. Without isolating prompt artifacts from target effects, the field risks misinterpreting these artifacts as simulated psychology.

To start addressing this gap, we develop a \textbf{framework that quantifies prompt artifacts in LM psychometric assessment.} Our design varies five prompt components: five semantically distinct baseline personas (assessment targets) and five semantically equivalent variants of each of four further components (persona wording, task instruction, item wording, and option symbol). We compare the response distributions by 1-Wasserstein distance and partition the total variation among the five components. Because only the baseline persona is construct-relevant, this separates target effects from prompt artifacts. We test this framework in a case study of 13 open-weight small language models (SLMs), where artifacts displace response distributions further than the baseline persona in most configurations.

\begin{figure*}
    \centering
    \includegraphics[width=\linewidth]{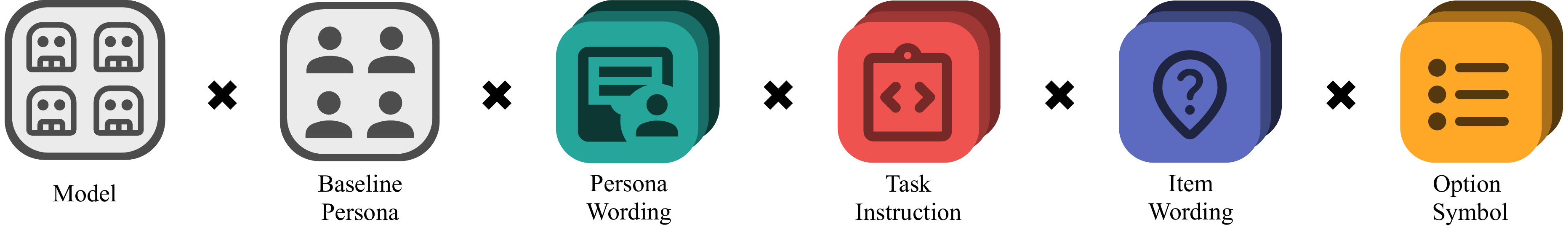}
    \caption{
        \textbf{Prompt Variant Matrix} \HYPRCBE: The Cartesian product of the semantically equivalent prompt variations (colored components). These four axes define the complete prompt space evaluated for each item.
    } 
    \label{fig:prompt.variant.matrix}
\end{figure*}

Our work makes two primary contributions to the evaluation and critique of LM psychometrics:

\noindent\textbf{A Variation Partitioning Framework:}
We introduce a diagnostic framework that separates prompt artifacts from target effects in LM-based survey experiments. Varying five prompt components over a balanced design with only one construct-relevant component (the baseline persona), we measure the Average Pairwise Wasserstein Distance (APWD) between response distributions and partition the variation among the five.

\noindent\textbf{Quantifying Artifact Dominance:} 
Applying this framework across 13 open-weight models (0.6B to 14B), we quantify an artifact dominance that raises concerns about the validity for the tested configurations. For most models and most items the artifact share of explained variation exceeds the target's variation, and scaling to 14B parameters narrows but does not close the gap. Because the artifact share frequently exceeds 50\%, the interpretation of these predictions is compromised well before any psychological claim is made.
\section{Related Work}
As in classical psychometrics, the measurement instrument in LM psychometrics is a prompt, built from text chosen by the researcher: task instruction, item wording, and response options, each can be formulated in multiple valid ways. However, unlike real psychometrics, the same prompt also defines the persona taking the test, meaning the prompt (and the model's weights) serves as both the measurement instrument and the respondent. The model conditions on all parts of the prompt, with nothing separating the component that carries the construct from the rest, making prompt sensitivity an inherent challenge in LM psychometrics.

\paragraph{Prompt Sensitivity} Research defines prompt sensitivity as a model's robustness against semantically equivalent rephrasings of the same prompt \cite{errica-etal-2025-wrong}. Those perturbations need not be semantic: arbitrary formatting choices, such as casing or spacing, shift predictions substantially \cite{sclar2023quantifying}. \citet{hua-etal-2025-flaw} counter that the literature's over-reliance on exact-match and log-likelihood metrics artificially inflates prompt sensitivity, masking the true robustness of LMs.

\paragraph{LM Psychometrics} The field of LM psychometrics splits into several subquestions. The most prominent of these asks how far model responses align with those of individuals \cite{ahnert2025survey} or of subpopulations \cite{pmlr-v202-santurkar23a, ahnert2025survey, munker2025a}. These studies find that alignment is partial, uneven across groups, and sensitive to how the group is specified. A second line examines whether the scores satisfy core psychometric criteria, such as validity and reliability \cite{tosato2026persistent, munker-2025-fingerprinting, lee2025llms}, and finds that they frequently do not. Combining both interests, a third line asks whether models reproduce human response biases \cite{rupprecht2025prompt, tjuatja2024llms}, finding that select models reproduce some of those biases.

\paragraph{Methodological Advances} Methodological research in LM psychometrics examines different approaches to prompt formulation \cite{lutz2025prompt}, as well as comparing different response generation methods \cite{ahnert2025survey, dominguez2024questioning}. This comparison matters, because the approaches are not neutral: \citet{wang2024my} show that a model's first token probabilities disagree with its own generated text answers, and \citet{rottger2024political} demonstrate that forced choice prompts, required to obtain valid responses, lead to constrained predictions that diverge from unconstrained ones, aligning with \citeauthor{hua-etal-2025-flaw}'s (\citeyear{hua-etal-2025-flaw}) argument that part of the measured sensitivity is an artifact of the readout itself.

\paragraph{Prompt Perturbations} Combining research on psychometrics, annotation, and multiple-choice questions with LMs, this research area tests model robustness to perturbations. Item reformulations and scale variants \cite{tosato2026persistent}, option symbol sets \cite{yang2025option}, prompt architecture \cite{10.1371/journal.pone.0319159}, response option design \cite{dominguez2024questioning}, and formatting choices \cite{sclar2023quantifying}, have each been shown to move responses, establishing that the exact phrasing of a prompt matters.

\paragraph{Research Gap} What no existing research establishes is \textbf{how much the perturbations matter relative to the construct the instrument is meant to measure}. Varying one axis and reporting the spread shows that a feature is influential, but it cannot say whether that influence is small or large compared to the intended psychological signal, because the signal is held fixed. This comparison requires manipulating construct-relevant and construct-irrelevant components in the same design and decomposing both on a common scale \cite{messick89}. We therefore construct such a design to quantify prompt artifacts for a given model $\times$ questionnaire combination. 
\section{Methods}
\label{sec:methods}

\subsection{Experimental Setup}

\paragraph{Models} 
We evaluate 13 instruction-tuned models (with reasoning deactivated) across four open-weight families: \texttt{Llama-3.1/3.2} \cite{grattafiori2024llama}, \texttt{Qwen3} \cite{yang2025qwen3}, \texttt{Ministral-3} \cite{liu2026ministral}, and \texttt{Gemma-4} \cite{gemmateam2026gemma4}.

\paragraph{Psychometric Inventories} 
We utilize the Big Five Inventory (BFI) \cite{zis-JohnSrivastava1999The} and the Short Dark Triad (SD3) \cite{jones2014introducing}, each using a 5-point Likert scale. Both are popular instruments belonging to the standard repertoire of current LM psychometrics.

\paragraph{Baseline Personas} 
To establish psychological baselines, we select five distinct personas from the Nemotron dataset, using their default texts \cite{nvidia/Nemotron-Personas-USA}. These five personas are the assessment targets, representing the only construct-relevant component in the design.

\subsection{Constructing the Prompt Variant Matrix \HYPRCBE}
\label{sec:methods:matrix}
To construct the Prompt Variant Matrix (Figure~\ref{fig:prompt.variant.matrix}), we use five variants for each component while preserving its semantic meaning. The goal of these variations is not to create an exhaustive grid search of all possible prompt combinations, but to measure the amount of variation within a contained set that remains within the reasonable bounds of what researchers might realistically design. Full variants are detailed in Appendix~\ref{sec:app:variants}.

\noindent\textbf{Persona Wording} \Persona 
We supplement each baseline persona by creating four variations. These variations systematically alter vocabulary and sentence order while preserving the underlying meaning, ensuring the target remains constant.

\noindent\textbf{Task Instruction} \Instruction 
We adapted our instructions from standard LM psychometric research. However, following \citeauthor{wang2024my}'s (\citeyear{wang2024my}) findings on first-token probability deviations, we augmented these baselines with formatting constraints. 

\noindent\textbf{Item Wording} \Question 
We utilize the item reformulations introduced by \citet{tosato2026persistent}. For each BFI and SD3 item, we test five lexical variants: the original questionnaire wording and four randomly sampled paraphrases from their validated dataset.

\noindent\textbf{Option Symbol} \OptionSymbol 
Based on prevalent formatting practices in the literature \cite{yang2025option}, we test five distinct ordinal option symbol sets: Arabic numerals (\texttt{1-5}), Roman numerals (\texttt{I-V}), uppercase letters (\texttt{A-E}), lowercase letters (\texttt{a-e}), and an arbitrary alphanumeric baseline (\texttt{x1-x5}).

\subsection{Metrics}

\subsubsection{Logit-Based Distribution Extraction}
We apply softmax ($t=1$) over the full vocabulary at the answer position and read the probability mass of the option tokens, computing the joint probability for multi-token variants as the product of their sequential token probabilities. By treating the output as a distribution, we capture the model's underlying uncertainty across the trait spectrum, avoiding the information loss caused by greedy text generation. 
\begin{figure*}[!t]
    \centering
    \includegraphics[width=\linewidth]{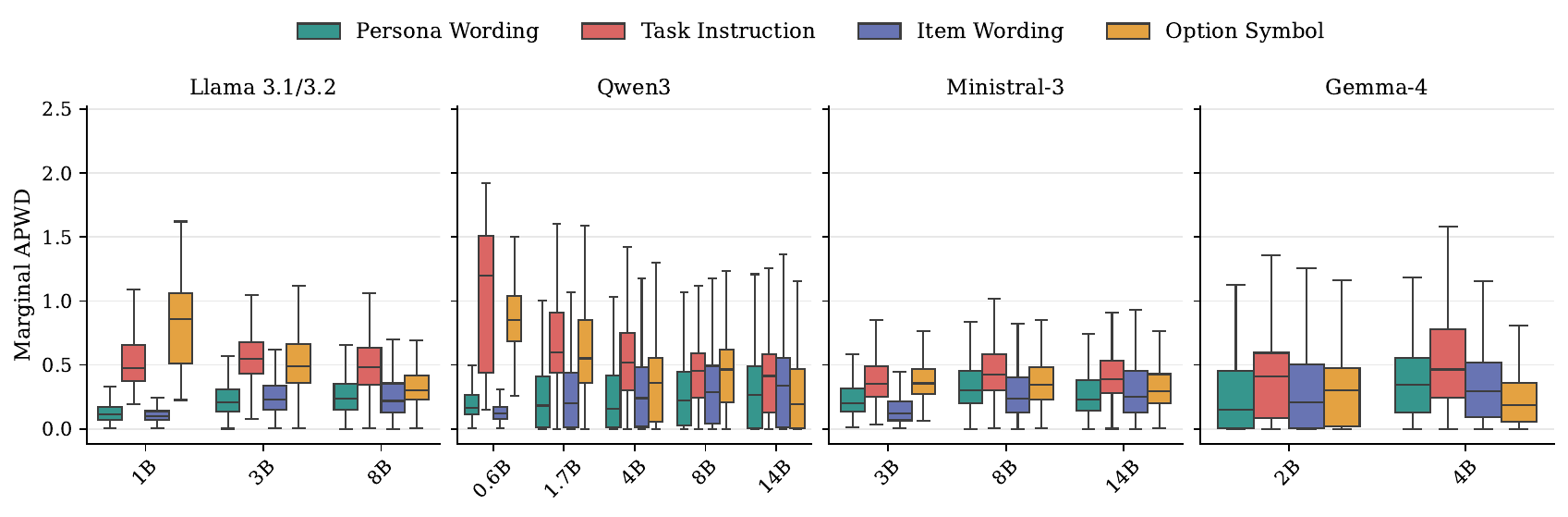}
    \caption{Marginal APWD for individual prompt components on the BFI dataset. Boxplots display the isolated displacement induced by altering a single component while holding the remaining components constant. SD3 results in Figure~\ref{fig:marginal_variance_sd3}.}
    \label{fig:marginal_variance}
\end{figure*}

\subsubsection{Displacement (APWD)} 
Because Likert responses are ordinal, standard probability metrics (e.g., Jensen-Shannon divergence) fail to penalize the ordinal magnitude of an error. Therefore, we quantify distributional shifts using the 1-Wasserstein distance ($W_1$), which calculates the minimum transport cost required to transform one distribution into another.

\paragraph{Total APWD} We compute the APWD across all $5^4$  combinations within a single Prompt Variant Matrix \HYPRCBE. This establishes an LM's geometric instability for a specific psychometric item and a specific persona. The reported score averages the APWDs from all 220 unique matrices (5 baseline personas $\times$ 44 BFI items).

\paragraph{Marginal APWD} To isolate the displacement caused by individual prompt components, we calculate the marginal APWD by varying only the target component while keeping the others fixed. Consequently, each marginal effect in a single model represents the APWDs across 27,500 isolated matrices (5 baseline personas $\times$ $5^3$ fixed variants $\times 44$ BFI items).

\subsubsection{Variation Partitioning}
We partition the total sum of squares, computed from the $W_1$ distances, among the five prompt components, using the distance-based decomposition that follows \citeauthor{anderson2014permutational}'s (\citeyear{anderson2014permutational}) approach. It operates directly on a distance matrix and requires no parametric assumption about the response distributions. Each component is a distance-based $R^2$ ($db\text{-}R^2$): the share of the total variation in response distributions that is explained by a prompt component rather than remaining within its levels (details in Appendix~\ref{sec:app:decomposition}). We analyze the artifact share of explained variation (ASV),
$$
\mathrm{ASV} = \frac{\sum_{k \in A} db\text{-}R^2_k}
                    {\sum_{k \in A} db\text{-}R^2_k + db\text{-}R^2_{\text{baseline}}}
$$
with $A$ representing the four construct-irrelevant components and $db\text{-}R^2_{\text{baseline}}$ the component of the baseline persona. Our claims concern the distribution of ASV across items rather than significance at any single item.

\section{Experiments}
\label{sec:experiments}

\subsection{Total Magnitude of Displacement}
\begin{table}[!b]
\small
\begin{tabular}{lcc}
\toprule
Model & BFI & SD3 \\
\midrule
Llama-3.2-1B & 0.815 ($\pm$ 0.039) & 0.803 ($\pm$ 0.059) \\
Llama-3.2-3B & 0.682 ($\pm$ 0.101) & 0.752 ($\pm$ 0.179) \\
Llama-3.1-8B & 0.583 ($\pm$ 0.145) & 0.724 ($\pm$ 0.178) \\
Ministral-3-3B & 0.504 ($\pm$ 0.111) & 0.516 ($\pm$ 0.074) \\
Ministral-3-8B & 0.638 ($\pm$ 0.157) & 0.759 ($\pm$ 0.148) \\
Ministral-3-14B & 0.592 ($\pm$ 0.185) & 0.740 ($\pm$ 0.184) \\
Qwen3-0.6B & 1.053 ($\pm$ 0.029) & 1.058 ($\pm$ 0.027) \\
Qwen3-1.7B & 0.718 ($\pm$ 0.170) & 0.874 ($\pm$ 0.166) \\
Qwen3-4B & 0.603 ($\pm$ 0.242) & 0.752 ($\pm$ 0.187) \\
Qwen3-8B & 0.626 ($\pm$ 0.187) & 0.784 ($\pm$ 0.196) \\
Qwen3-14B & 0.579 ($\pm$ 0.286) & 0.728 ($\pm$ 0.219) \\
Gemma-4-E2B & 0.542 ($\pm$ 0.257) & 0.815 ($\pm$ 0.260) \\
Gemma-4-E4B & 0.643 ($\pm$ 0.238) & 0.877 ($\pm$ 0.192) \\
\bottomrule
\end{tabular}
\caption{Mean (standard deviation) APWD across all Prompt Variant Matrices for the BFI and SD3 datasets. Higher values indicate greater measurement error.}
\label{tab.total_apwd}
\end{table}
Table~\ref{tab.total_apwd} reports the APWD. The APWD values between 0.5 and 1.0 mean that, averaged over pairs of semantically equivalent prompts, the response distributions vary by roughly half to one full ordinal step. We observe distinct sensitivity towards the datasets: for the BFI, the models tend to exhibit lower displacement than on the SD3, suggesting weaker SD3 trait representation in the Nemotron personas or within the model. Given this instability, SD3 results are deferred to Appendix~\ref{sec:app:sd3}. Further, the data reveals scaling differences between the models. While Llama and Qwen stabilize at larger scales, Ministral and Gemma become more volatile. 

\subsection{Marginal Displacement of Components}

Across almost all evaluated architectures, task instructions\Instruction and option symbols\OptionSymbol drive higher marginal displacement than wording (Figure~\ref{fig:marginal_variance}). Furthermore, vulnerabilities carry distinct architectural signatures, such as Qwen's sensitivity to instructions\Instruction versus Llama's vulnerability to option symbol variants\OptionSymbol. 

We also observe intra-family divergence. For instance, Qwen3-14B develops an above-average sensitivity to item wording variations\Question absent in its smaller counterparts. We observe no unified pattern of robustness within, let alone across, model families. Finally, we must contextualize this volatility: while moderate instability indicates active attention to a prompt component, the near-zero marginal displacement observed in small models (e.g., Llama-3.2-1B) does not indicate robustness. Instead, it indicates a failure to process the component, as we will show in the next section.

\subsection{Variation Partitioning}
\begin{figure}[!htbp]
    \centering
    \includegraphics[width=\linewidth]{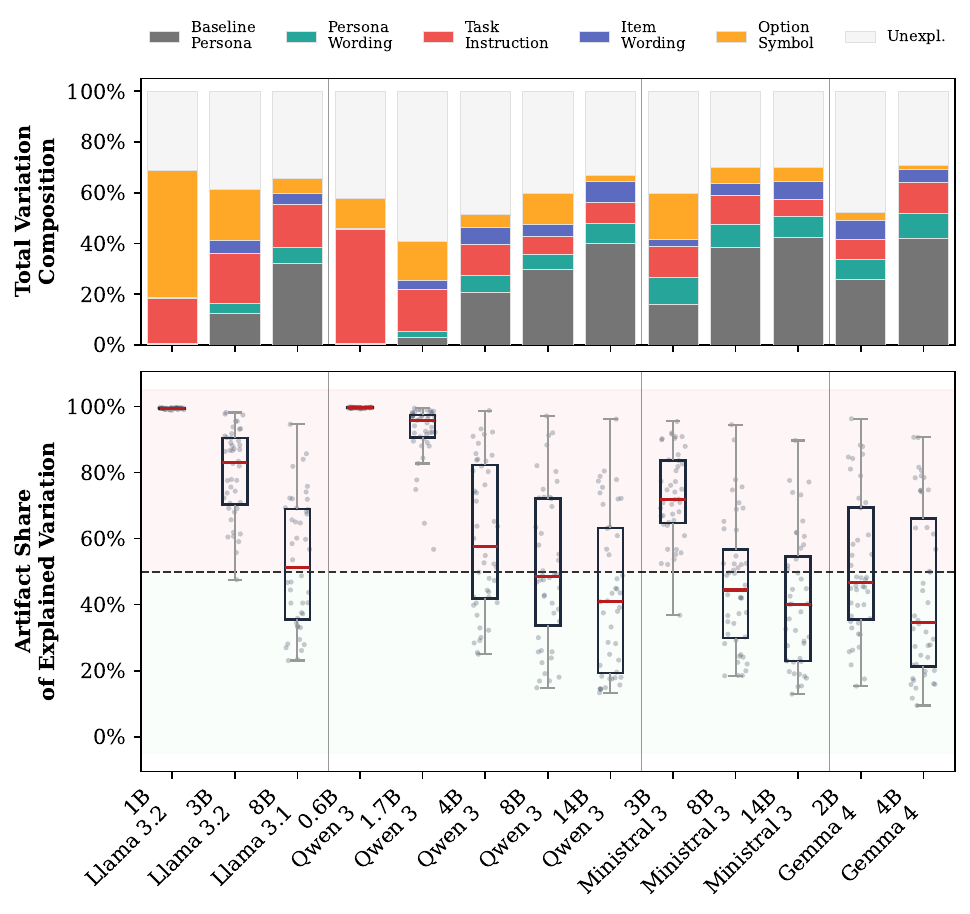}
    \caption{
        Distance-based variation partitioning on the BFI dataset. (Top) Mean proportion of total variation explained by the baseline persona versus other prompt components. (Bottom) The artifact share of explained variation (per questionnaire item). SD3 results in Figure~\ref{fig:variance_decomposition_sd3}.}
    \label{fig:variance_decomposition}
\end{figure}

Figure~\ref{fig:variance_decomposition} (Top) reports the mean variation composition across all 44 BFI items. In sub-billion models, the intended psychological signal is close to zero (e.g., the small Qwen model collapses due to instruction-induced distributional shifts\Instruction, the small Llama due to option symbol variants\OptionSymbol). As models scale toward 14B, the target effect strengthens. However, these average improvements mask severe item-level failures. 

Figure~\ref{fig:variance_decomposition} (Bottom) demonstrates that even at 14B, some items yield an ASV $>$ 50\%, indicating that prompt artifacts have a bigger impact than the baseline personas. Because the artifact-explained variation remains this large, none of the evaluated models demonstrated sufficient robustness for the psychometric configurations tested here. Measuring up to the 14B scale consistently risks reflecting prompt artifacts rather than intended signal. Finally, large parts of the variation remain unexplained due to interactions between prompt components, further compounding the problem.
\section{Conclusion \& Future Work}
We developed the Prompt Variant Matrix and distance-based variation partitioning and found that prompt artifacts remain substantial across tested open-weight models up to 14B parameters, demonstrating a lack of robustness for the tested questionnaire $\times$ model combinations. One possible explanation is that instruction tuning introduces strong format compliance that competes with the injected persona for control of the answer position \cite{wang2024my, rottger2024political}. 

While this dominance raises concerns about the validity of SLMs for psychometric use under current prompting paradigms, our framework offers three paths forward. First, it serves as a diagnostic tool: researchers can use this design to identify the prompt components and specific variants that steer a model away from its ``default'' distribution. Second, it quantifies prompt artifacts for a given questionnaire $\times$ model combination, which downstream designs can report and account for. Third (untested here), the framework enables an accuracy assessment. Averaging a persona's response distributions across the Prompt Variant Matrix\HYPRCBE marginalizes out the prompt artifacts, leaving artifact-adjusted predictions for that persona. Scored against human respondents completing the same questionnaire, this would test how accurate the remaining signal is, as our results only establish how much of the measurement stems from prompt artifacts. 
\section*{Limitations}

While our Prompt Variant Matrix and distance-based variation partitioning provide a diagnostic framework for quantifying prompt artifacts, this study operates within several methodological boundaries:

\paragraph{Non-Exhaustive Parameter Space} The structural axes of the Prompt Variant Matrix represent a targeted, not exhaustive, sample of potential prompts. We bounded our experimental design to instruction constraints, option symbols, and wording. We did not evaluate variations in structural placement (e.g., shifting the baseline persona from the system prompt to the user prompt) or ordinal directionality (e.g., reversing the order of answer options), which prior literature \cite{zeng2024evaluating, wang2024large} suggests could induce further distributional shifts.

\paragraph{Sensitivity to Variant Selection \& Trait Density} The magnitude of the measured variation is dependent on our specific variant configurations. While we anchored our prompts in published designs to keep the variant set representative of current practice, the framework remains sensitive to variants that diverge too sharply. Furthermore, the baseline Nemotron personas may lack the latent trait density necessary to measurably anchor complex Big Five or SD3 dimensions. This weak initial trait representation could exacerbate the severe artifact dominance we observed.

\paragraph{Semantic Similarity Claim}
Our design treats the prompt variants as semantically equivalent; a premise that cannot be verified exhaustively and that prompt sensitivity research adopts under comparable conditions \cite{errica-etal-2025-wrong}. We intentionally scope ``semantic equivalence'' as ``what researchers might realistically design'' (compare Section~\ref{sec:methods:matrix}), and draw our component variants from published research. Only the persona wordings\Persona are authored by us. Our results show that where equivalence is contestable the stakes are also lowest: item wordings\Question, which come from a validated reformulation set \cite{tosato2026persistent}, and the persona wording variants\Persona, we wrote ourselves, produce the smallest displacement for most model $\times$ item combinations we tested. Option symbols\OptionSymbol, which carry no construct content, and the task instructions\Instruction, whose equivalence is close to definitional, account for most of the variation in most of the tested configurations. Lastly, exact equivalence would in any case be self-defeating as a target, as variants a model cannot distinguish would produce identical response distributions and no measurable displacement at all. 

\paragraph{First-Token Probability Divergence}
We read responses from the option token probabilities at the answer position, despite documented divergence between first-token distribution and open-text answers \cite{wang2024my, rottger2024political, ahnert2025survey}. We would still consider this the standard readout in LM psychometrics, which aligns with the scope defined in the previous paragraph: the prompt artifacts we report are those a researcher following current practice encounters. While our results are subject to this error, our framework does not depend on this choice. It requires a distribution over the ordinal response options per item, which repeated sampling or other approaches supply equally well. Whether artifact dominance persists under those readouts is an open question this design can answer without modification, and one our case study cannot settle.

\paragraph{Restriction to Small-Scale Open Models} While the observed artifact dominance restricts the operational use of SLMs in psychometric environments, we intentionally restricted our evaluation to open-weight architectures to ensure transparent, reproducible access to the response distributions our readout requires. Consequently, we cannot definitively conclude whether closed-weight, proprietary models (e.g., ChatGPT, Claude) successfully overcome prompt sensitivity.

\paragraph{Absence of Human Baselines} We evaluated structural instability purely \textit{in silico}. It is a well-documented phenomenon in classical psychometrics \cite{krosnick1987evaluation} that human subjects also exhibit response-order bias when item wording or scale formats change. Because we did not benchmark against a parallel human control group, we cannot precisely quantify the divergence between expected human methodological variance and the total variation observed in the SLMs.

\paragraph{Exclusion of Interaction Effects} Our decomposition quantifies main-effect contributions only. While complex interaction effects between models, psychological constructs, and prompt components undoubtedly exist, our data reveal that individual SLMs possess highly idiosyncratic prompt sensitivities. Attempting to model more generalized interaction effects may overcomplicate the analysis and obscure the fundamental finding: that the effects of these artifacts are already sufficient to overpower the intended psychological signal.
\section*{Ethical Considerations}
The primary ethical consideration arising from our findings relates to how researchers and practitioners interpret LM-generated survey responses. Our contribution is that this doesn't need to rest on assumptions: the framework quantifies prompt artifacts for a given model $\times$ questionnaire combination before any psychological claim is made, so that researchers can report it, select prompt configurations with a lower artifact share, or establish that a setup is not yet suitable for the inference they intend. 

\paragraph{Reproducibility Constraints} Finally, our findings underscore the necessity of methodological transparency. To prevent the misinterpretation of prompt artifacts as simulated traits, researchers must publicly release their exact prompt templates and parameters. Without this transparency, claims about LM capabilities in computational social science cannot be validated.
\section*{Acknowledgements}
This work was conducted within the Ecological Validation of Artificial Simulacra (EVAS) Research Cooperative (\url{https://evas-research.github.io/}). This work is supported by EMO-SCI (project number 569156112) funded by Deutsche Forschungsgemeinschaft (DFG, German Research Foundation); and KuKI (project number 16KIS2515), funded by the Bundesministerium für Forschung, Technologie und Raumfahrt (BMFTR, German Federal Ministry of Research, Technology and Space).
\bibliography{custom}
\appendix

\section{Reproducibility and Code Availability}
\label{app:reproducibility}

All experimental procedures, statistical analyses, and visualization tools are implemented using Python 3.13+ with standard scientific computing libraries (NumPy \cite{harris2020array}, SciPy \cite{virtanen2020scipy}, pandas \cite{mckinney-proc-scipy-2010}, seaborn \cite{waskom2021seaborn}). The Transformers \cite{wolf-etal-2020-transformers} library was used to run the models on one NVIDIA L40S. All code implementations, experimental configurations, processed datasets, results and analysis notebooks that can be made publicly available, are accessible via our repository (\url{https://github.com/nsschw/unsampled-truth}).

\section{Generative AI Assistance}
During coding, we utilized GitHub Copilot's inline coding suggestions. Additionally, we employed Gemini 3 Pro for documentation and minor coding tasks, all of which were manually verified. During the writing process, we relied on Gemini 3 Pro, DeepL Translate, and DeepL Write for stylistic refinements and proofreading.

\section{Variants}
\label{sec:app:variants}
\begin{table*}[htbp]
\centering
\small
\renewcommand{\arraystretch}{1.3} 
\begin{tabularx}{\textwidth}{X}
\midrule
\multicolumn{1}{c}{\textbf{Persona Variations\Persona (Baseline: Antonia Crockett)}} \\
\midrule
You are Antonia Crockett, 28, a tech-savvy DIY enthusiast who blends a love for street photography, yoga, and community mentorship into a balanced life, though you sometimes procrastinate on financial reviews and can't resist a late-night coffee-fueled coding session. -- \textit{inspired by} \citet{nvidia/Nemotron-Personas-USA}\\ 
You are 28 and tech-oriented, mixing your DIY passion with photography, yoga, and mentorship. While your life is balanced, you tend to delay financial tasks and often find yourself in coffee-fueled coding sessions late at night. -- \textit{self-created}\\ 
At 28 years of age, you live your life between yoga mats, your camera, and your handcrafted projects. While you balance your community through your mentorship, your bank balance and its reviews get left out for late-night coffee and some lines of code. -- \textit{self-created}\\ 
At 28, you balance the physical (yoga), the visual (photography), and the tangible (handcrafted art). While you're a pillar for your community, your administrative life and coding projects only come alive under the glow of a desk lamp and a late-night brew. -- \textit{self-created}\\ 
You've built a life around wellness and art, prioritizing human connection above all else. However, this leaves your financial struggles to be dealt with in the dark, where you balance the heavy weight of economic pressure against the logic of your code. -- \textit{self-created}\\ 
\midrule
\multicolumn{1}{c}{\textbf{Instruction Variations\Instruction}} \\
\midrule
You are participating in a psychological evaluation. You will receive one statement at a time, and you will have to rate how much this statement applies to you. Choose a \{option\_name\} from \{range\} as your rating: \{option\_mapping\}. Your ENTIRE response must be ONLY the \{option\_name\} between \{range\} that represents your rating. Do not include any other text, punctuation, or explanation - your response should be only a single \{option\_name\}. If unsure, make your best guess. Don't overthink it.\textbackslash nHere is the statement: '\{item\}' -- \textit{inspired by} \citet{tosato2026persistent}\\ 
You are participating in a personality traits assessment and are provided with statements of characteristics that may or may not apply to you. For example, do you agree that you are someone who likes to spend time with others? Your task is to select the best fitting option from the following set to indicate the extent to which you agree or disagree with that statement:\textbackslash n\{option\_mapping\}\textbackslash nChoose the option that aligns best with your personality without overthinking and answer the following statement: '\{item\}'\textbackslash nPlease refrain from adding additional considerations to your choices and only answer with the \{option\_name\} -- \textit{inspired by} \citet{tagarelli2025}\\ 
Given a statement of you: '\{item\}'\textbackslash nPlease choose from the following options to identify how accurately this statement describes you. Answer only with a \{option\_name\} of the following options: \{option\_mapping\}  -- \textit{inspired by} \citet{lee2025llms}\\ 
For the following task, respond in a way that matches your description. Considering the statement, please indicate the extent to which you agree or disagree answering only with a \{option\_name\} from the scale \{range\} (where \{option\_mapping\}): '\{item\}' -- \textit{inspired by} \citet{zou2024can}\\ 
Rate how much you agree with the statement below on a scale from \{range\} using the following options:\textbackslash n\{option\_mapping\}\textbackslash nReply with exactly one \{option\_name\}.\textbackslash nStatement: '\{item\}' -- \textit{self-created}\\ 
\midrule
\multicolumn{1}{c}{\textbf{Item Variations\Question (Example: BFI - Extroversion)}} \\
\midrule
Is talkative -- \textit{original from} \cite{zis-JohnSrivastava1999The} \\
Has a lot to say  -- \textit{re-used from} \citet{tosato2026persistent} \\
Has a tendency to talk extensively  -- \textit{re-used from} \citet{tosato2026persistent}\\
Often talks a great deal  -- \textit{re-used from} \citet{tosato2026persistent}\\
Is often chatting -- \textit{re-used from} \citet{tosato2026persistent}\\
\midrule
\multicolumn{1}{c}{\textbf{Option Symbol Variations\OptionSymbol}} \\
\midrule
Arabic Numerals (1--5) \cite{yang2025option}\\
Uppercase Letters (A--E) \cite{yang2025option}\\
Roman Numerals (I--V) \cite{yang2025option}\\
Alphanumeric Baseline (x1--x5) \cite{yang2025option}\\
Lowercase Letters (a--e) -- \textit{self-created}\\
\bottomrule
\end{tabularx}
\caption{A complete demonstrator slice of the Prompt Variant Matrix\HYPRCBE. This table demonstrates all variations of the four prompt components for an example case. The Cartesian product of these specific rows yields the 625 discrete prompts evaluated for this single baseline persona and BFI item combination. Variables enclosed in brackets (e.g., \texttt{\{option\_mapping\}}) denote injection points (e.g., mapping the option symbol to the semantic anchors: Strongly Disagree, Disagree, Neutral, Agree, Strongly Agree).}
\label{tab:appendix_demonstrator_matrix}
\end{table*}
Table~\ref{tab:appendix_demonstrator_matrix} shows for one persona and item the variants we used. All other configurations can be found in the repository.

\section{Variation Decomposition}\label{sec:app:decomposition}We partition variation using distance-based sums of squares \citep{anderson2014permutational}, a non-parametric method operating directly on interpoint Wasserstein distances, $d_{ij}=W_1(P_i,P_j)$. For the $N=3{,}125$ balanced prompt cells per model~$\times$~item, total variation is:$$SS_{\mathrm{T}} = \frac{1}{N}\sum_{i<j} d_{ij}^{2}$$For a component $k$ with $L$ levels of size $n_{\ell}$, the \emph{within-level} variation restricts this sum to cell pairs sharing level $\ell$:$$SS_{\mathrm{W}}(k) = \sum_{\ell=1}^{L}\frac{1}{n_{\ell}} \sum_{i<j\,\in\,\ell} d_{ij}^{2}$$The marginal share of total variation explained by $k$ is its distance-based $R^2$:$$db\text{-}R^{2}_{k} = \frac{SS_{\mathrm{T}} - SS_{\mathrm{W}}(k)}{SS_{\mathrm{T}}}$$Because components are evaluated marginally, their shares need not sum to the total modelled variation.

\section{Additional Results}
\begin{figure*}[!htbp]
    \centering
    \includegraphics[width=\linewidth]{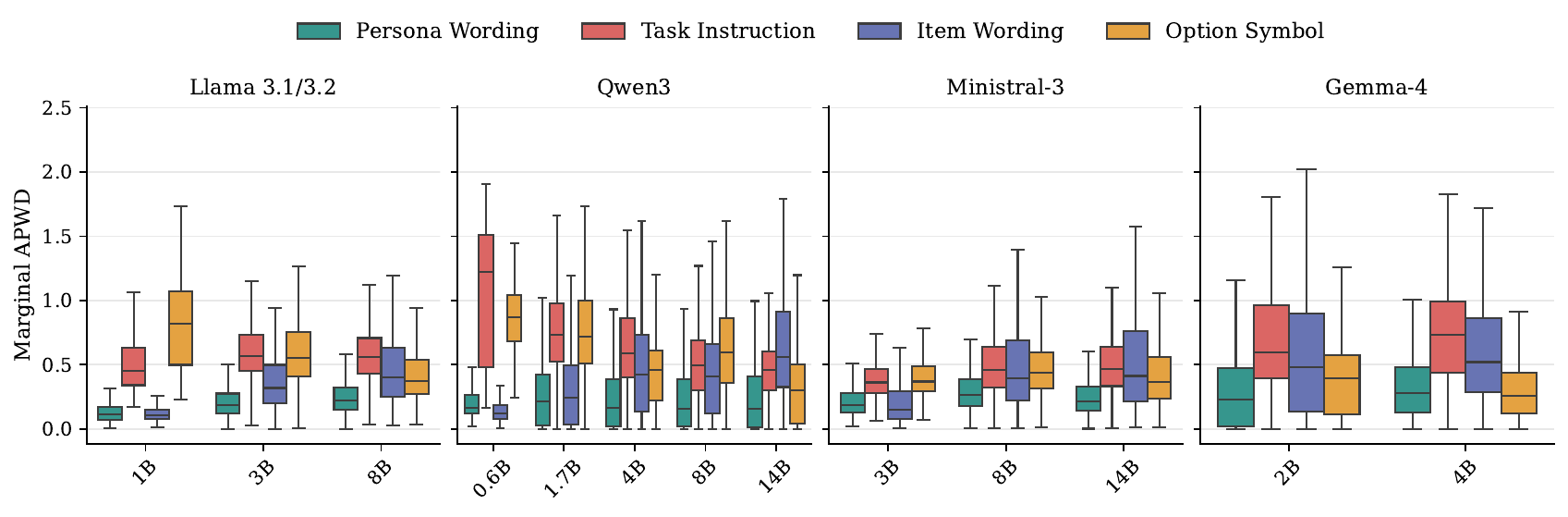}
    \caption{Marginal APWD for individual prompt components on the SD3 dataset. Boxplots display the isolated displacement induced by altering a single component while holding the remaining components constant.}
    \label{fig:marginal_variance_sd3}
\end{figure*}

\subsection{Off-Option Mass}
\begin{table*}[!htpb]
\centering
\small
\renewcommand{\arraystretch}{1.1}
\begin{tabular}{lcccc}
\toprule
& \multicolumn{2}{c}{\textbf{BFI}} & \multicolumn{2}{c}{\textbf{SD3}} \\
\cmidrule(lr){2-3} \cmidrule(lr){4-5}
\textbf{Model} & \textbf{APWD} & \textbf{OOM} & \textbf{APWD} & \textbf{OOM} \\
\midrule
Llama-3.2-1B & 0.815 ($\pm$ 0.039) & 0.149 ($\pm$ 0.018) & 0.803 ($\pm$ 0.059) & 0.122 ($\pm$ 0.018) \\
Llama-3.2-3B & 0.682 ($\pm$ 0.101) & 0.113 ($\pm$ 0.050) & 0.752 ($\pm$ 0.179) & 0.129 ($\pm$ 0.102) \\
Llama-3.1-8B & 0.583 ($\pm$ 0.145) & 0.129 ($\pm$ 0.043) & 0.724 ($\pm$ 0.178) & 0.135 ($\pm$ 0.057) \\
Ministral-3-3B & 0.504 ($\pm$ 0.111) & 0.181 ($\pm$ 0.021) & 0.516 ($\pm$ 0.074) & 0.184 ($\pm$ 0.021) \\
Ministral-3-8B & 0.638 ($\pm$ 0.157) & 0.335 ($\pm$ 0.029) & 0.759 ($\pm$ 0.148) & 0.361 ($\pm$ 0.035) \\
Ministral-3-14B & 0.592 ($\pm$ 0.185) & 0.313 ($\pm$ 0.034) & 0.740 ($\pm$ 0.184) & 0.318 ($\pm$ 0.028) \\
Qwen3-0.6B & 1.053 ($\pm$ 0.029) & 0.095 ($\pm$ 0.007) & 1.058 ($\pm$ 0.027) & 0.092 ($\pm$ 0.007) \\
Qwen3-1.7B & 0.718 ($\pm$ 0.170) & 0.101 ($\pm$ 0.028) & 0.874 ($\pm$ 0.166) & 0.107 ($\pm$ 0.027) \\
Qwen3-4B & 0.603 ($\pm$ 0.242) & 0.163 ($\pm$ 0.036) & 0.752 ($\pm$ 0.187) & 0.115 ($\pm$ 0.028) \\
Qwen3-8B & 0.626 ($\pm$ 0.187) & 0.001 ($\pm$ 0.001) & 0.784 ($\pm$ 0.196) & 0.001 ($\pm$ 0.006) \\
Qwen3-14B & 0.579 ($\pm$ 0.286) & 0.022 ($\pm$ 0.013) & 0.728 ($\pm$ 0.219) & 0.011 ($\pm$ 0.009) \\
Gemma-4-E2B & 0.542 ($\pm$ 0.257) & 0.116 ($\pm$ 0.023) & 0.815 ($\pm$ 0.260) & 0.100 ($\pm$ 0.023) \\
Gemma-4-E4B & 0.643 ($\pm$ 0.238) & 0.145 ($\pm$ 0.033) & 0.877 ($\pm$ 0.192) & 0.144 ($\pm$ 0.051) \\
\bottomrule
\end{tabular}
\caption{Mean (Standard Deviation) Average Pairwise Wasserstein Distance (APWD) and off-option mass (OOM). The OOM values describe the share of the model's predictions allocated to non-target tokens, representing a failure to adhere to prompt formatting constraints.}
\label{tab:apwd_ood_per_dataset}
\end{table*}
Table \ref{tab:apwd_ood_per_dataset} reports the off-option mass (OOM) alongside the APWD scores. Across most architectures, models allocate between 10\% and 15\% of their probability mass to non-target tokens. To compute the APWD, the remaining probability mass over the target ordinal symbols is subsequently normalized to one.

The data reveals distinct, family-specific scaling patterns regarding format compliance. The Ministral-3 family exhibits an inverse scaling trend, where OOM leakage expands from $\sim$18\% at 3B to over 30\% at the 8B and 14B scales. Conversely, the larger Qwen3 architectures (8B and 14B) eliminate OOM mass ($<2.2\%$), enforcing near-perfect formatting constraints. Notably, despite Qwen3-8B's and 14B's compliance in token generation, their APWD scores remain high, demonstrating that eliminating formatting leakage does not remove artifact variation in the response distributions. This also supports the understanding that the effects we find are not a confound of our prediction readout.

\subsection{SD3 Results}
\label{sec:app:sd3}
\begin{figure}[!b]
    \centering
    \includegraphics[width=\linewidth]{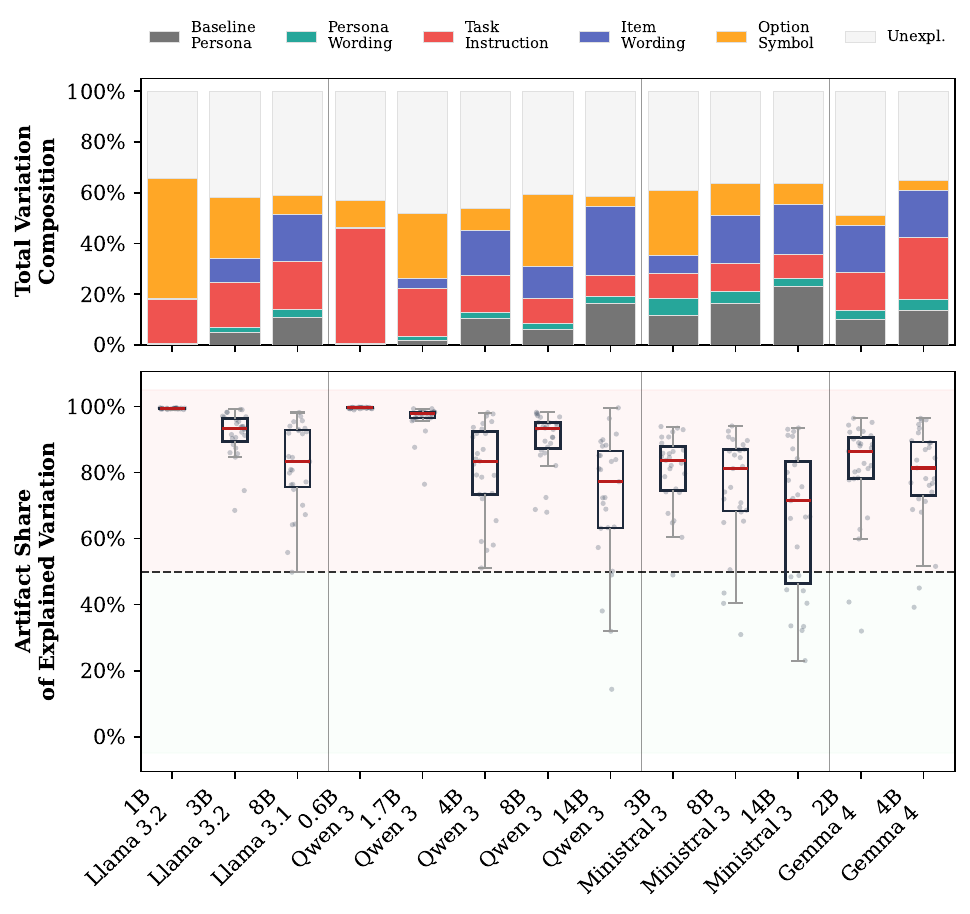}
    \caption{Distance-based variation partitioning on the SD3 dataset. (Top) Mean proportion of total variation explained by the baseline persona versus other prompt components. (Bottom) The artifact share of explained variation (per questionnaire item).}
    \label{fig:variance_decomposition_sd3}
\end{figure}

As established in Table~\ref{tab.total_apwd}, the SD3 dataset results in higher absolute instability across all evaluated architectures compared to the BFI. Figure \ref{fig:variance_decomposition_sd3} details the distance-based variation partition for these items, revealing a distinct degradation of the intended psychological signal. Figure \ref{fig:marginal_variance_sd3} confirms that the marginal effects of individual components, specifically task instructions\Instruction and option symbols\OptionSymbol, mirror the vulnerabilities observed in the BFI dataset but operate at a higher magnitude. While the BFI dataset demonstrated artifact dominance, the SD3 results exhibit even stronger ASV. For the majority of models, the median ASV shifts into the 80\%--100\% range, indicating that prompt artifacts dictate the predictions. Furthermore, because the OOM leakage rates remain similar across both datasets (Table \ref{tab:apwd_ood_per_dataset}), this artifact dominance occurs under equally strict format compliance, highlighting the risk of analyzing results with personas lacking sufficient depth, as the models maintain strict formatting compliance even as the target effects collapse.

\end{document}